%% file: main.tex
\documentclass{article}
\usepackage{iclr2027_conference,times}
\usepackage[T1]{fontenc}
\usepackage{amsmath,amssymb,graphicx,booktabs,array,longtable,multirow,tabularx}
\usepackage{xcolor,enumitem}
\usepackage[font=small]{caption}
\definecolor{SLDCiteBlue}{HTML}{2D85F0}
\definecolor{SLDLinkBlue}{HTML}{00308F}
\usepackage[pagebackref=true]{hyperref}
\hypersetup{
  colorlinks=true,
  citecolor=SLDCiteBlue,
  linkcolor=SLDLinkBlue,
  urlcolor=SLDCiteBlue,
  breaklinks=true,
  pdfborder={0 0 0}
}

\usepackage{url}
\input{numbers}

\newcommand{\bench}{SLDBench v2}
\newcommand{\nmse}{\operatorname{NMSE}}

\newcommand{\figref}[1]{Figure~\ref{#1}}
\newcommand{\tabref}[1]{Table~\ref{#1}}
\newcommand{\appref}[1]{Appendix~\ref{#1}}


\title{SLDBench v2: Active Scaling Law Discovery}
\input{authors}
\date{}

\iclrfinalcopy

\makeatletter
\renewcommand{\headrulewidth}{0pt}
\newcommand{\makefirstpage}{%
  \begingroup
    \setlength{\parindent}{0pt}%
    \normalfont
    \vspace*{-14pt}%
    {\centering
      {\fontsize{19.5}{23}\selectfont\bfseries\@title\par}%
      \vspace{13pt}%
      {\fontsize{10.5}{13}\selectfont
        Lingbei Meng\textsuperscript{1,*}\quad
        Rui Huang\textsuperscript{2,*}\quad
        Zhirong Shen\textsuperscript{3}\quad
        Jinxin Liu\textsuperscript{3}\par
        \vspace{2pt}%
        Lingxiao Jing\textsuperscript{3}\quad
        Haorui Chen\textsuperscript{4}\quad
        Yaosong Du\textsuperscript{3}\quad
        Haowei Lin\textsuperscript{5,\dag}\par}%
      \vspace{8pt}%
      {\fontsize{9}{11.5}\selectfont\color{black!80}%
        \textsuperscript{1}The Chinese University of Hong Kong, Shenzhen\quad
        \textsuperscript{2}The University of Hong Kong\\[1pt]%
        \textsuperscript{3}University of Electronic Science and Technology of China\quad
        \textsuperscript{4}Tsinghua University\quad
        \textsuperscript{5}Peking University\par}%
      \vspace{5pt}%
      {\fontsize{8.5}{11}\selectfont\ttfamily\color{black!80}%
        250010166@slai.edu.cn\quad huang\_rui@std.uestc.edu.cn\\[1pt]%
        2024080907011@std.uestc.edu.cn\quad linhaowei@pku.edu.cn\par}%
      \vspace{4pt}%
      {\fontsize{8.5}{10.5}\selectfont\color{black!80}%
        \textsuperscript{*}Equal contribution\quad
        \textsuperscript{\dag}Corresponding author\par}%
      \par}%
    \vspace{12pt}%
  \endgroup
}
\makeatother

\renewcommand{\maketitle}{\makefirstpage}

\renewenvironment{abstract}{%
  {\centering\fontsize{11}{13}\selectfont\scshape\color{black}Abstract\par}%
  \vspace{8pt}%
  \begin{list}{}{%
      \setlength{\leftmargin}{0.85cm}%
      \setlength{\rightmargin}{0.85cm}%
      \setlength{\listparindent}{0pt}%
      \setlength{\itemindent}{0pt}%
      \setlength{\topsep}{0pt}%
      \setlength{\parsep}{0pt}%
      \setlength{\itemsep}{0pt}}%
    \item\relax
}{%
  \end{list}%
  \vspace{6pt}%
}

\begin{document}
\maketitle
\input{sections/00_abstract}
\input{sections/01_introduction}
\input{sections/02_interface}
\input{sections/04_instantiation}
\input{sections/06_related_work}
\input{sections/07_limitations}
\label{main-text-end}
\input{sections/10_statements}

\bibliography{sections/references}
\bibliographystyle{iclr2027_conference}
\appendix
\input{sections/appendix}
\end{document}

%% file: numbers.tex
\newcommand{\SLDOriginalScenarios}{8}
\newcommand{\SLDStagingScenarios}{127}
\newcommand{\SLDStagingObservations}{3,376,938}
\newcommand{\SLDActiveEligibleScenarios}{60}
\newcommand{\SLDBenchmarkScenarios}{30}
\newcommand{\SLDDifficultyThreshold}{0.8}
\newcommand{\SLDBenchmarkObservations}{339,092}
\newcommand{\SLDBenchmarkPoolObservations}{266,183}
\newcommand{\SLDBenchmarkTestObservations}{72,909}
\newcommand{\SLDBenchmarkSources}{14}
\newcommand{\SLDEvaluatedSystems}{8}
\newcommand{\SLDExplicitPriceScenarios}{18}
\newcommand{\SLDProxyPriceScenarios}{12}
\newcommand{\SLDMinActivePool}{40}
\newcommand{\SLDMinPriceSpread}{20}
\newcommand{\SLDMinPriceLevels}{8}
\newcommand{\SLDReportedCostCoverage}{2}
\newcommand{\SLDBenchmarkReportedFLOPs}{1.09\times 10^{23}}
\newcommand{\SLDTokenizationUnits}{85}
\newcommand{\SLDTokenizationFLOPs}{2.70\times 10^{22}}
\newcommand{\SLDProteinUnits}{196}
\newcommand{\SLDProteinFLOPs}{8.24\times 10^{22}}
\newcommand{\SLDOWTUnits}{15}
\newcommand{\SLDOWTFLOPs}{1.55\times 10^{18}}

%% file: authors.tex
\author{%
  Lingbei Meng$^{1,*}$ \quad Rui Huang$^{2,*}$ \quad Zhirong Shen$^{3}$ \quad Jinxin Liu$^{3}$\\
  \bfseries Lingxiao Jing$^{3}$ \quad Haorui Chen$^{4}$ \quad Yaosong Du$^{3}$ \quad Haowei Lin$^{5,\dagger}$\\
  {\normalfont\small $^{1}$The Chinese University of Hong Kong, Shenzhen\quad $^{2}$The University of Hong Kong}\\
  {\normalfont\small $^{3}$University of Electronic Science and Technology of China\quad $^{4}$Tsinghua University\quad $^{5}$Peking University}\\
  {\normalfont\small\texttt{250010166@slai.edu.cn, huang\_rui@std.uestc.edu.cn}}\\
  {\normalfont\small\texttt{2024080907011@std.uestc.edu.cn, linhaowei@pku.edu.cn}}\\
  {\normalfont\small $^{*}$Equal contribution\quad $^{\dagger}$Corresponding author}
}

%% file: sections/00_abstract.tex
\begin{abstract}
Scaling laws predict large-model performance from small experiments, but finding a reliable law itself requires costly experiments. SLDBench tests whether language-model agents can discover scaling laws when all experimental results are given upfront. We introduce \bench{} with two extensions. First, we expand the corpus from \SLDOriginalScenarios{} to \SLDStagingScenarios{} scenarios spanning language-model pretraining, downstream performance, and vision--language transfer, and select \SLDBenchmarkScenarios{} challenging scenarios on which a reference power-law fit reaches $R^2\le\SLDDifficultyThreshold$; they represent at least $\SLDBenchmarkReportedFLOPs$ FLOPs of source-reported training compute. Second, we introduce \emph{active scaling law discovery}: an agent chooses which experiments to run, pays each one's recorded cost (e.g., training FLOPs), and revises its law as outcomes arrive, and is scored on the same hidden extrapolation targets as in the passive setting. Across \SLDEvaluatedSystems{} frontier agent systems, median $R^2$ drops from 0.72--0.75 (passive) to 0.24--0.57 (active), and prediction error rises 1.5--4.7$\times$, depending on the system. Agents favor cheap experiments, spending 8--44\% of the full-pool cost on average; low-cost acquisition can yield strong predictions, but cheap purchases also leave experimental conditions needed for extrapolation unobserved. Experimental planning thus remains a key open challenge for agents that assist model development in the prevailing AI-for-AI study.
Project page: \url{https://lingbei-meng.github.io/SLDBench-v2/}.
\end{abstract}

%% file: sections/01_introduction.tex
\section{Introduction}
\label{sec:introduction}

Scaling laws guide foundation-model development by predicting how performance changes with model size, training data, and computation \citep{kaplan2020,hoffmann2022}. Their practical value lies in extrapolation: experiments at smaller scales inform decisions about larger training runs, such as how to split compute between model size and data. A reliable law, however, requires experiments of its own. Researchers must choose which configurations to measure and how much compute to spend before the predictions can guide investment. Efficient scaling law discovery therefore depends on both the quality of a law and the cost of the evidence behind it.

Recent work shows that language-model agents can contribute to this process. SLDBench evaluates whether agents can discover scaling laws from completed experiments across \SLDOriginalScenarios{} scenarios, while SLDAgent searches over law expressions and fitting procedures to improve extrapolation beyond established human-derived laws on the evaluated tasks \citep{lin2026}. This formulation preserves an essential feature of scientific discovery: the predictive objective is specified, but the mathematical form of the solution is open-ended. The experimental evidence, however, is supplied upfront, and its cost plays no role. Success in this \emph{passive} setting leaves open whether an agent can discover an accurate law when it must also decide which experiments are worth conducting.

Choosing those experiments is closely coupled to choosing the law. Candidate laws can agree on observed small-scale behavior yet diverge sharply at larger scales. Additional inexpensive measurements may leave this ambiguity unresolved, whereas an experiment at a different scale or parameter combination may help distinguish the candidates. The agent's current hypotheses guide which measurements it seeks next, and a new observation may require revising those hypotheses. The agent must therefore reason about what its evidence can establish, which additional evidence would improve extrapolation, and whether that improvement justifies the experimental cost. Neither the number of acquired measurements nor their goodness of fit alone captures this capability.

We introduce \bench, extending SLDBench in two complementary directions. First, we collect a corpus of \SLDStagingScenarios{} scenarios comprising \SLDStagingObservations{} prediction records from public experimental data. We construct paired Passive and Active settings for \SLDActiveEligibleScenarios{} eligible scenarios, then retain the \SLDBenchmarkScenarios{} for which the reference power-law fit scores $R^2\leq\SLDDifficultyThreshold$ in both settings. This screening focuses evaluation on scenarios that remain challenging for a standard law and numerical fitting procedure. The resulting benchmark spans \SLDBenchmarkSources{} sources, including language-model pretraining, downstream performance, and vision--language transfer, with \SLDBenchmarkObservations{} observations. The source-reported training compute of only \SLDReportedCostCoverage{} of these sources already amounts to $\SLDBenchmarkReportedFLOPs$ FLOPs (\appref{app:corpus-cost}).

\begin{figure}[!t]
\centering\includegraphics[width=\linewidth]{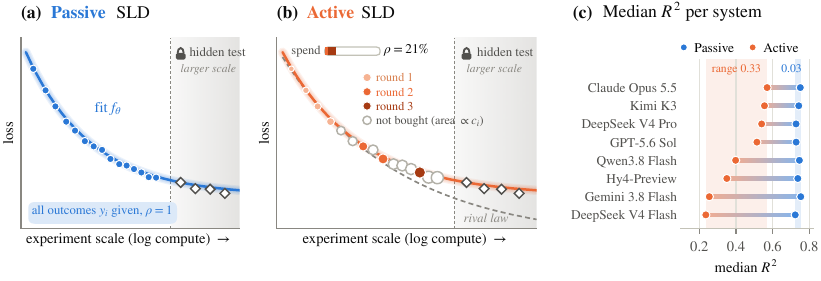}
\caption{Passive and active scaling law discovery. (a)~Passive SLD supplies the outcomes of all candidate experiments; the agent fits a law that must extrapolate to hidden larger-scale runs. (b)~Active SLD supplies only a priced catalogue. The agent buys batches of experiments over rounds (filled markers, shaded by round; area grows with price), refits its law after each round, and stops; it is scored on the same hidden runs, and its spending $\rho$ is reported. A rival law (dashed) that fits the cheap points diverges at the test scale. (c)~Median $R^2$ of the \SLDEvaluatedSystems{} agent systems on the \SLDBenchmarkScenarios{} tasks: nearly identical in the Passive setting and widely spread in the Active one; shaded bands mark the range of median scores in each setting (\tabref{tab:systems}).}
\label{fig:protocol}
\end{figure}

Second, we extend the benchmark to \emph{active scaling law discovery} (Active SLD), in which agents allocate experimental resources while constructing a predictive law (\figref{fig:protocol}). An agent receives candidate configurations and their costs, selects experiments, observes their outcomes, and uses the evidence to revise both its law and its next choices. It must also decide when to stop acquiring evidence and submit its law. The form of the law and its fitting procedure remain open-ended. Active SLD thus extends the objective from accurate extrapolation on a fixed dataset to jointly balancing extrapolation quality and experimental cost, such as the FLOPs needed for new measurements.

We implement Active SLD through historical experiment replay: selecting an experiment reveals its recorded outcome and charges its price. This allows decisions about expensive experiments to be evaluated repeatedly without rerunning any training. Each scenario also retains the original passive setting, which provides all candidate outcomes upfront. Both settings use the same hidden extrapolation targets, with predictive quality and experimental cost reported separately.

We evaluate \SLDEvaluatedSystems{} frontier agent systems with 720 runs in each setting (Section~\ref{sec:experiments}). Three findings stand out. First, the Passive setting no longer separates these systems, whose median $R^2$ lies between 0.723 and 0.753, whereas the Active setting spreads them from 0.235 to 0.570, and the Passive ranking only weakly predicts the Active one. Second, the Active loss is a heavier failure tail rather than a uniform shift, and it does not come from sequential access: runs that buy the full pool match their Passive counterparts, so the loss arises from what is bought. Third, agents favor cheap experiments, which can leave groups and parameter combinations unobserved; nevertheless, Claude Opus 5.5 leads in Active at 4.01\% median cost. Non-agent rules point the same way: under a fixed law, sampling each group's price range beats covering every group with its cheapest experiments. By bringing experimental acquisition into scaling law discovery, \bench{} provides a testbed for AI-for-AI research in which agents decide how to spend the compute needed to develop future models.

%% file: sections/02_interface.tex
\section{SLDBench v2}
\label{sec:benchmark}

\begin{figure}[t]
\centering\includegraphics[width=\linewidth]{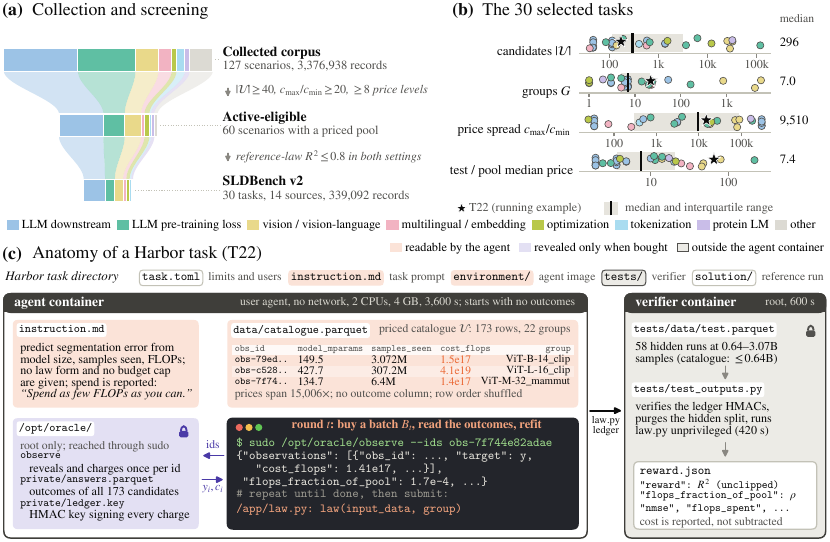}
\caption{Construction and anatomy of \bench. (a)~The three-stage funnel from the collected corpus to the benchmark, colored by research domain; domains are assigned by keyword rules. (b)~The 30 selected tasks. Each dot is a task and the star is the running example T22; the bar and the number on the right give the median, and the band the interquartile range; the last row divides the median test price by the median candidate price. (c)~The Harbor task directory and the two containers of one task. The agent reads the instruction and the priced catalogue, buys outcomes through a root-only oracle that signs every charge into a ledger, and submits an executable law; a separate verifier checks the ledger and scores the law on hidden larger-scale runs. File names, commands, and limits are those of the released task.}
\label{fig:dataset}
\end{figure}

This section formalizes passive and active scaling law discovery (SLD), describes how the \bench{} tasks are collected and screened, and specifies the execution environment and metrics. \figref{fig:dataset} summarizes the construction and follows one task from catalogue to score.

\paragraph{General formulation.}
\label{sec:formulation}
Following SLDBench~\citep{lin2026}, an SLD task is a set of experimental records $\mathcal D=\{(\mathbf x_i,g_i,y_i)\}_{i=1}^{m}$, where each record consists of three components:
\begin{enumerate}[leftmargin=2em,nosep,label=(\arabic*)]
\item a \emph{feature vector} $\mathbf x_i\in\mathbb R^{d}$ of experimental choices, such as parameter count $N$, training tokens or samples $D$, training compute, learning rate, batch size, or other hyper-parameters;
\item a \emph{control index} $g_i\in\mathcal G$ identifying the context that is held fixed within a single experimental series, such as a model architecture, a training objective, or a family of models;
\item a \emph{target} $y_i\in\mathbb R$, such as a validation loss, an error rate, or a downstream accuracy.
\end{enumerate}
The goal is a symbolic law $f$ with group-specific parameters $\theta_g$ such that $\hat y=f(\mathbf x;\theta_g)$ predicts a hidden test set $\mathcal T$ of larger-scale experiments from the same task. The functional form of $f$ is shared across groups, whereas its coefficients may differ. The form itself is not given: for pre-training loss, $L(N,D)=E_g+A_gN^{-\alpha_g}+B_gD^{-\beta_g}$~\citep{hoffmann2022} is one hypothesis, and the agent decides which variables enter the law, which parameters are shared across groups, and how the law is fitted. SLDBench supplies the complete set $\mathcal D$ upfront; we call this setting \emph{Passive SLD}.

\paragraph{Active formulation.}
\label{sec:interaction}
Active SLD withholds the targets and exposes a priced catalogue $\mathcal U=\{(i,\mathbf x_i,g_i,c_i)\}_{i=1}^{n}$, where $c_i>0$ is the cost of running experiment $i$, typically its training FLOPs. The agent starts with $S_0=\emptyset$. At round $t$ it holds the revealed history $H_t=\{(i,y_i):i\in S_t\}$ and either stops or buys a new batch $B_t\subseteq\mathcal U\setminus S_t$, whose targets are then revealed:
\begin{equation}
S_{t+1}=S_t\cup B_t,\qquad C(S)=\sum_{i\in S}c_i,\qquad \rho(S)=\frac{C(S)}{C(\mathcal U)}\in[0,1].
\label{eq:cost}
\end{equation}
At its stopping round $\tau$ the agent submits a law fitted to $H_\tau$. A solution therefore couples three decisions, an acquisition policy, a stopping rule, and a law, and is evaluated by the pair $\bigl(\nmse_{\mathcal T}(f),\rho(S_\tau)\bigr)$: how well the law extrapolates, and what fraction of the full-pool cost was spent to obtain it. Passive SLD is the special case $S_0=\mathcal U$ and $\rho=1$. The coupling is what makes the problem hard. Two candidate laws that agree on the purchased points can diverge at the test scale, and only some purchases, often the expensive ones, can separate them; which purchases are informative in turn depends on the laws still under consideration. Active SLD therefore differs from cost-aware fitting of a supplied law family~\citep{li2026active}, where the hypothesis space is fixed, and from pool-based active learning, where test inputs are drawn from the same range as the pool: here the test experiments lie beyond the scale of every candidate and cannot be bought.

\paragraph{Example.}
\figref{fig:dataset}c shows task T22, from the LAION comparison of CLIP and MaMMUT training (\appref{app:inventory}). Its features are model size $N$, samples seen $D$, and training compute; its target is ADE20K segmentation error, $1-\mathrm{mIoU}$; and its 22 groups are 11 vision-transformer architectures, each trained with either objective. The catalogue offers 173 runs whose prices span a 15{,}006-fold range, while the 58 test runs use 0.64--3.07B samples and no candidate exceeds 0.64B. A natural hypothesis is $\hat y_g(D)=E_g+A_gD^{-\alpha_g}$, but within a group larger $D$ costs more. The agent must trade breadth against depth: two cheap points in every group cannot identify a per-group exponent, whereas full series for a few groups identify the curve shape only if exponents transfer across groups. Section~\ref{sec:value} analyzes a run that makes the second choice without testing it.

\paragraph{Datasets.}
\label{sec:tasks}
We build \bench{} in three stages (\figref{fig:dataset}a). (i)~\emph{Collection.} We gather \SLDStagingScenarios{} scenarios with \SLDStagingObservations{} prediction records from public papers, repositories, and training logs, covering language-model pre-training loss and downstream accuracy, vision and vision--language transfer, multilingual embeddings, optimization hyper-parameters, tokenization, and protein language models. Each scenario records its source, features, target, groups, and extrapolation split; structural curation merges duplicate feature--group keys and removes uninformative features, and provenance and exclusion reasons are retained (\appref{app:taskdetails}). (ii)~\emph{Active eligibility.} A scenario supports a meaningful purchase decision only if every candidate has a positive price, $|\mathcal U|\geq\SLDMinActivePool$, $c_{\max}/c_{\min}\geq\SLDMinPriceSpread$, at least \SLDMinPriceLevels{} distinct prices occur, and, when test prices are recoverable, the median test price exceeds the median candidate price; \SLDActiveEligibleScenarios{} scenarios qualify. (iii)~\emph{Difficulty screen.} We fit a fixed reference, a shared additive power law estimated by bounded multi-start least squares (\appref{app:screening}), and keep a scenario only if the reference scores $R^2\leq\SLDDifficultyThreshold$ in both settings; no agent score enters the screen. The result is \SLDBenchmarkScenarios{} tasks from \SLDBenchmarkSources{} sources with \SLDBenchmarkObservations{} records, of which \SLDBenchmarkPoolObservations{} are candidates and \SLDBenchmarkTestObservations{} are test points. The tasks are heterogeneous (\figref{fig:dataset}b): catalogues contain 40 to 118{,}480 candidates in 1 to 5{,}575 groups, the median price spread is 9{,}510-fold, and in every task the median test experiment costs more than the median candidate (1.9 to 218 times; median 7.4).

\paragraph{Execution environment and metrics.}
Each task is packaged in the Harbor format~\citep{harbor2026}, the standard task format and evaluation harness of agentic benchmarks such as Terminal-Bench~2.0~\citep{merrill2026terminal}. A Harbor task is a self-contained directory holding a natural-language instruction, an environment image, a verifier test suite, and a configuration of resources and time limits, so any command-line agent can be evaluated on it without adaptation. A \bench{} task runs as a pair of containers (\figref{fig:dataset}c). The agent container runs as an unprivileged user without network access, with 2 CPUs, 4\,GB of memory, and a 3{,}600\,s limit. It holds the instruction and the catalogue, which has no target column. The outcomes reside in a root-only directory reachable only through \texttt{sudo /opt/oracle/observe --ids}; the tool returns the targets and prices of the requested candidates, charges each distinct candidate once, and appends an HMAC-signed entry to a spend ledger; repeated purchases are free. The instruction names the target and features, requires one functional form for all groups, gives no law form and no budget cap, and states that spending is reported alongside accuracy. The agent submits \texttt{/app/law.py}, which defines \texttt{law(input\_data, group)}, and a written explanation. A separate verifier container checks the ledger signatures, runs the law as an unprivileged user under a 420\,s limit, and scores its predictions on the hidden test set by the normalized mean squared error
\begin{equation}
\nmse_{\mathcal T}(f)=\frac{\sum_{i\in\mathcal T}\bigl(y_i-f(\mathbf x_i;\theta_{g_i})\bigr)^2}{\sum_{i\in\mathcal T}(y_i-\bar y_{\mathcal T})^2},\qquad R^2=1-\nmse_{\mathcal T}(f).
\label{eq:score}
\end{equation}
The reward is the unclipped $R^2$, or zero if the predictions are not finite. $\rho(S_\tau)$ is reported next to it and never subtracted, because any fixed exchange rate between error and FLOPs would encode one practitioner's preference. A negative $R^2$ means worse than predicting the test mean. The two settings of a task share one test file, so their $\nmse$ ratio equals their MSE ratio. Replay reveals recorded outcomes instead of launching training jobs, so an episode that buys $10^{20}$ FLOPs costs only the agent's inference. Prices are source-reported FLOPs, documented estimates such as $6ND$, or non-FLOP proxies (\appref{app:corpus-cost}); agent inference and wall-clock time are not charged.

\paragraph{Specialty of \bench.}
Existing formulations fix one of the two unknowns. SLDBench and symbolic regression~\citep{udrescu2020,cranmer2023} fix the data and ask for the law; Bayesian experimental design~\citep{huan2013} and SL$^2$~\citep{li2026active} fix the law family and ask for the data. \bench{} asks for both, while fixing the target, the test set, and the prices and logging every purchase. Error and cost are therefore comparable across systems, and each episode is an auditable record of which experiments an agent chose and in what order (Section~\ref{sec:experiments}).

%% file: sections/04_instantiation.tex
\section{Experiments and Analysis}
\label{sec:experiments}

\subsection{Experimental setup}
\label{sec:setup}

\paragraph{Systems and runs.}
We evaluate \SLDEvaluatedSystems{} agent systems, each a frontier model in a command-line agent harness (\tabref{tab:systems}): six models in Claude Code, GPT-5.6 Sol in Codex, and Gemini 3.8 Flash in Gemini CLI. Each system solves every task three times in each setting under the protocol of Section~\ref{sec:benchmark}, which gives 90 runs per system and setting and 720 runs per setting. Implementation details are given in \appref{app:protocol}.

\paragraph{Metrics.}
Quality and cost are reported separately. For quality we report the median $R^2$ over runs with a finite score and threshold shares, the percentage of all scheduled runs whose $R^2$ exceeds 0, 0.5, or 0.9, so that a missing score counts as a failure. To compare the settings on a scale-free footing we use the error multiplier $(1-R^2_{\mathrm A})/(1-R^2_{\mathrm P})$ between the Active and Passive runs of the same system, task, and repeat. Because tasks from one source are not independent, we take its median within each of the 14 sources and then across sources, with 95\% intervals from 10{,}000 source resamples (\appref{app:score-rules}). Cost is the full-pool cost fraction $\rho(S_\tau)$ of \eqref{eq:cost}. Appendix~\ref{app:best-at-three} additionally reports best@3 $R^2$, taking the highest of three scores for each task before aggregating across tasks.

\paragraph{Baselines.}
Two kinds of reference place the agents in context. (i)~The \emph{reference law} of the difficulty screen, fitted to the full pool. (ii)~\emph{Non-agent acquisition rules}: the same reference law fitted to candidates chosen without looking at any outcome, at budgets from 0.5\% to 100\% of the pool cost (\appref{app:baselines}). \emph{Cheapest-first} buys in ascending price; \emph{random} buys in a random order (three random seeds); \emph{cheapest-per-group} visits the groups in round-robin order and takes each group's next-cheapest candidate; \emph{price-spread} visits the groups in the same way but samples each group's price range coarse to fine, starting from its cheapest and most expensive candidates.

\subsection{Main results}
\label{sec:results}

\begin{table}[!t]
\centering\footnotesize
\caption{Prediction quality and acquisition cost, sorted by median Active $R^2$. $R^2$ columns are medians over finite scores; ${>}.5$ is the percentage of the 90 scheduled runs per system and setting with $R^2>0.5$; $\Delta R^2$ is the Active minus the Passive median. Cost is the Active acquisition cost as a percentage of the full-pool cost, and full is the percentage of Active runs that bought every candidate. Bold marks the best agent value of each quality column. $^\dagger$The reference law of the difficulty screen (\appref{app:screening}), fitted once to all candidates, where the two settings coincide; the screen keeps only tasks on which it scores $R^2\leq\SLDDifficultyThreshold$.}
\label{tab:systems}
\setlength{\tabcolsep}{2.4pt}
\begin{tabular*}{\linewidth}{@{\extracolsep{\fill}}llrrrrrrrr@{}}
\toprule
& & \multicolumn{2}{c}{Passive} & \multicolumn{3}{c}{Active} & \multicolumn{3}{c}{Active cost $\rho$ (\%)}\\
\cmidrule(lr){3-4}\cmidrule(lr){5-7}\cmidrule(l){8-10}
Model & Harness & $R^2$ & ${>}.5$ & $R^2$ & ${>}.5$ & $\Delta R^2$ & mean & median & full\\
\midrule
Claude Opus 5.5 & Claude Code & 0.751 & \textbf{80.0} & \textbf{0.570} & \textbf{55.6} & \textbf{$-$0.181} & 7.8 & 4.01 & 1.1 \\
Kimi K3 & Claude Code & 0.742 & 72.2 & 0.555 & 45.6 & $-$0.187 & 32.0 & 17.75 & 10.0 \\
DeepSeek V4 Pro & Claude Code & 0.728 & 64.4 & 0.540 & 51.1 & $-$0.188 & 34.8 & 21.51 & 7.8 \\
GPT-5.6 Sol & Codex & 0.728 & 70.0 & 0.514 & 51.1 & $-$0.214 & 43.8 & 35.20 & 11.1 \\
Qwen3.8 Flash & Claude Code & 0.746 & 74.4 & 0.398 & 42.2 & $-$0.348 & 15.8 & 5.35 & 2.2 \\
Hy4-Preview & Claude Code & 0.737 & \textbf{80.0} & 0.350 & 42.2 & $-$0.387 & 9.1 & 3.44 & 0.0 \\
Gemini 3.8 Flash & Gemini CLI & \textbf{0.753} & 70.0 & 0.255 & 42.2 & $-$0.499 & 25.9 & 7.78 & 15.6 \\
DeepSeek V4 Flash & Claude Code & 0.723 & 66.7 & 0.235 & 41.1 & $-$0.488 & 39.3 & 17.59 & 26.7 \\
\midrule
Reference law$^\dagger$ & -- & 0.455 & 50.0 & 0.455 & 50.0 & -- & 100.0 & 100.00 & 100.0 \\
\bottomrule
\end{tabular*}
\end{table}

\begin{figure}[!t]
\centering\includegraphics[width=\linewidth]{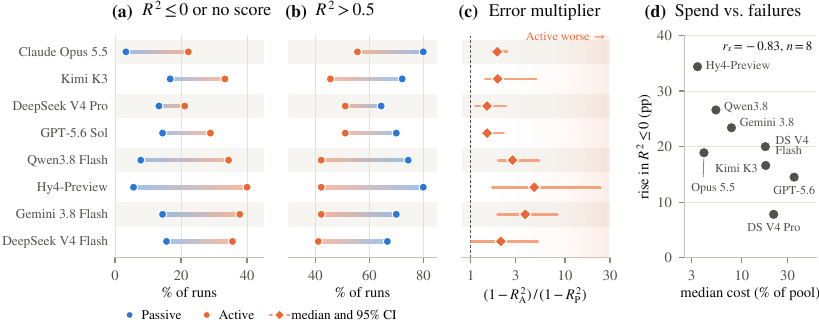}
\caption{Where the Active loss comes from. (a)~Percentage of scheduled runs with $R^2\leq0$ or no score and (b)~with $R^2>0.5$, in the Passive (blue) and Active (orange) settings. (c)~Active/Passive error multiplier of each system, aggregated within and then across the 14 sources; bars are 95\% source-bootstrap intervals, and the dashed line marks equal error. (d)~Median Active cost of each system against the Passive-to-Active rise in its failure share of panel~(a); $r_s$ is the Spearman rank correlation over the \SLDEvaluatedSystems{} systems.}
\label{fig:outcomes}
\end{figure}

\paragraph{Active SLD separates systems that Passive SLD does not.}
\label{sec:gap}
The Passive medians range from 0.723 to 0.753, a range of 0.030 (\tabref{tab:systems}). When the systems must buy their data, the medians span 0.235 to 0.570, a range of 0.334, and the Passive ranking only weakly predicts the Active one (Spearman $r_s=0.21$, $n=\SLDEvaluatedSystems$). Gemini 3.8 Flash has the best Passive median, but the seventh-best Active median and the largest drop, $-0.499$; Claude Opus 5.5, second in the Passive setting, leads in the Active one, followed by Kimi K3. Every system loses quality: its source-aggregated error multiplier lies between 1.49 and 4.71, with every 95\% interval above one (\figref{fig:outcomes}c). The full-pool reference law gives an absolute scale. In the Passive setting every system exceeds it by 0.27 to 0.30; in the Active setting the four strongest systems do, at median costs of 4--35\% of the pool.

\paragraph{The loss is a heavier failure tail.}
Threshold shares locate the loss (\figref{fig:outcomes}a,b). Pooled over systems, the share of runs with $R^2\leq0$ or no score rises from 11.4\% to 31.7\%, the share above 0.5 falls from 72.2\% to 46.4\%, and the share above 0.9 falls from 19.3\% to 8.2\% (\tabref{tab:coverage-v4}). Between 45.6\% and 55.6\% of the four strongest systems' runs still exceed 0.5, yet 21.1\% to 40.0\% of each system's Active runs predict worse than the test mean or produce no score.

\paragraph{Sequential access is not the obstacle.}
An Active run that buys every candidate sees the same data as the Passive run through a different interface. Among the 67 such pairs, Active is better by more than 0.01 in $R^2$ in 38.8\%, Passive in 29.9\%, and the remaining 31.3\% differ by at most 0.01; the median difference is near zero. These runs are self-selected, but their parity shows that the interface itself imposes no penalty. The Active loss therefore stems from what is bought.

\paragraph{Acquisition cost alone does not determine prediction quality.}
Lower spending is associated with a larger rise in failure share (\figref{fig:outcomes}d). Across the \SLDEvaluatedSystems{} systems, median cost is negatively correlated with the rise in failure share ($r_s=-0.83$) and the fall in the share above 0.5 ($r_s=-0.71$). Cost alone does not determine quality: Claude Opus 5.5 achieves the highest Active median, 0.570, at a median cost of 4.01\% of the pool. Hy4-Preview and Qwen3.8 Flash also spend little, at medians of 3.44\% and 5.35\%, but raise their failure shares most, by 34.4 and 26.7 percentage points. At higher spending, DeepSeek V4 Flash and Kimi K3 have almost the same median cost, 17.59\% and 17.75\%, yet Active medians of 0.235 and 0.555; DeepSeek V4 Flash spends bimodally, with 22.2\% of its runs below 1\% of the pool and 27.8\% above 90\% (\tabref{tab:system-behavior}).

\subsection{What agents buy}
\label{sec:spending}

\begin{figure}[!t]
\centering\includegraphics[width=\linewidth]{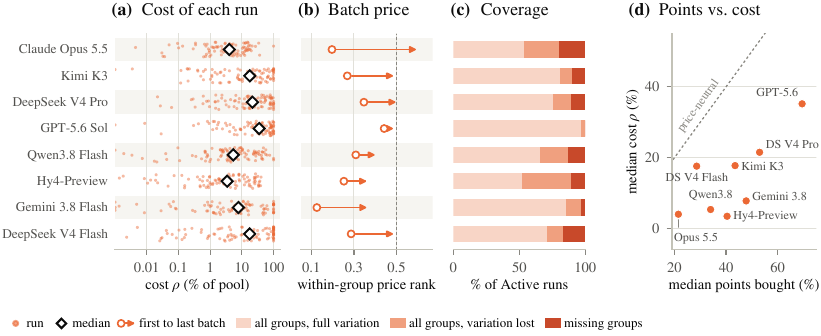}
\caption{What agents buy. (a)~Active cost of every run as a percentage of the full-pool cost (dots) and the per-system median (diamonds). (b)~Median within-group price rank of the first (circle) and last (arrowhead) batch of multi-batch runs; 0 is the cheapest candidate of a group and 1 the most expensive. (c)~Runs that observe every group with the catalogue's full independent parameter variation, every group with reduced variation, or not every group. (d)~Median share of candidates bought against median share of the full-pool cost, each computed separately; points below the diagonal pay less than the catalogue's mean price per candidate.}
\label{fig:behavior}
\end{figure}

\paragraph{Cheap first, expensive later.}
Agents prefer cheap candidates. For every system the median share of candidates bought exceeds the median share of cost (\figref{fig:behavior}d), and the median ratio of the mean price paid to the catalogue mean ranges from 0.18 to 0.90 (\tabref{tab:buying-v4}); Hy4-Preview buys 40.4\% of the candidates for 3.44\% of the cost. Within one system, run costs span more than three orders of magnitude (\figref{fig:behavior}a), so a system is better described by a spending distribution than by a budget. The order of purchases follows a common pattern: for every system the median first batch lies in the cheaper half of each group's price range and the median last batch lies higher in that range (\figref{fig:behavior}b). The typical episode is thus cheap exploration followed by costlier confirmation.

\paragraph{Cheap acquisition leaves conditions unobserved.}
Although 89.3\% of Active runs observe every group, 16.7\% do so but lose independent parameter variation that the catalogue offers, and 10.7\% omit at least one group (\figref{fig:behavior}c). Even when every group is observed, independent variation is lost in 36.7\% of Hy4-Preview runs, 26.7\% of Claude Opus 5.5 runs, and 21.1\% of Qwen3.8 Flash runs, against 3.3\% for GPT-5.6 Sol, the highest spender. Such omissions surface at test time. One Hy4-Preview run on OpenWebText buys 51 points but never observes batch size 64 or 128 at learning rate $10^{-4}$, both present in the test set, and scores $R^2=-83.3$; three of the 24 OpenWebText runs miss such a combination (\appref{app:cases}).

\paragraph{Outcomes vary across repeats.}
The same system on the same task often reaches qualitatively different results. Depending on the system, the sign of $R^2$ changes across the three repeats on 20.0\% to 46.7\% of tasks, and GPT-5.6 Sol, the highest spender, has a more than tenfold error range on 33.3\% of tasks (\tabref{tab:system-behavior}). Because acquisition is sequential, an early batch that suggests a wrong law form can plausibly steer the rest of an episode. Per-system medians over three repeats are thus coarse, and differences between adjacent rows of \tabref{tab:systems} should not be over-read.

\subsection{What the purchased data are worth}
\label{sec:value}

\begin{figure}[!t]
\centering\includegraphics[width=\linewidth]{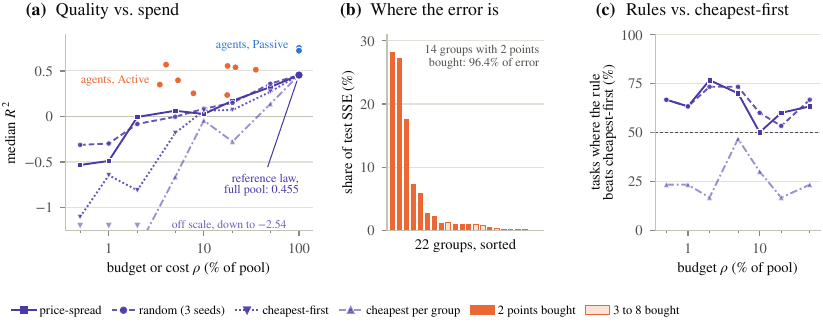}
\caption{What the purchased data are worth. (a)~Median $R^2$ over the 30 tasks when the fixed reference law is fitted to candidates chosen by four non-agent rules at budget $\rho$ (lines), with the agents' Passive medians at $\rho=100\%$ and their Active medians at their median cost; each agent marker pairs two separate medians. (b)~Share of the test squared error per group in the DeepSeek V4 Pro run on T22, sorted; dark bars are groups with two purchased points. (c)~Share of the 30 tasks on which a rule gives the reference law a higher $R^2$ than cheapest-first selection at the same budget; ties count as not higher, and the dashed line marks parity.}
\label{fig:value}
\end{figure}

\paragraph{A fixed law needs far more data than the agents buy.}
We fit the reference law to candidates chosen by the four rules (\figref{fig:value}a). Its median $R^2$ falls steeply with budget, from 0.455 with the full pool to at most 0.36 at $\rho=50\%$ and at most 0.085 at $\rho=10\%$, and it is negative under every rule at $\rho\leq2\%$. The Active medians of the four strongest systems, 0.514--0.570 at median costs of 4--35\%, lie above every rule's curve and above the full-pool reference. The comparison conflates acquisition with law choice and favors the agents by construction, since the difficulty screen keeps tasks on which the reference law fails. It shows only that the agents extract more predictive value from partial data than a fixed law does, not that they select their data efficiently.

\paragraph{Spanning each group's price range matters more than covering every group.}
The rules also show which property of a cheap purchase set matters most (\figref{fig:value}c). Random and price-spread selection beat cheapest-first on 63\% to 77\% of tasks at $\rho\leq5\%$. Cheapest-per-group, which reaches every group early but only through its cheapest candidates, beats cheapest-first on at most 46.7\% of tasks at any budget, and its median $R^2$ is the lowest of the four rules at every budget below the full pool. Price-spread visits the groups in the same order and differs only in sampling each group's price range, and at $\rho=2\%$ it raises the median $R^2$ from $-1.456$ (cheapest-per-group) to $-0.005$. Under a fixed law, cheap points in every group add little information; variation in price within a group, and hence in scale, is what identifies the law, and it is exactly what costs the most.

\paragraph{Error concentrates where an assumption goes untested.}
Returning to T22, DeepSeek V4 Pro buys eight points in each of four cheap groups to learn the curve shape and two early-scale points in each of 14 other groups, then transfers a shared exponent, $\hat y_g=E_g+A_gD^{-0.05}$, to all groups. The run buys 78 points for 4.407\% of the pool and scores $R^2=-10.839$. The 14 sparsely sampled groups contribute 96.4\% of the test squared error and the three largest contributors 73.2\%, with extrapolation up to 1{,}000 times the largest purchased sample count (\figref{fig:value}b). The failure is not a shortage of points but an assumption the purchases never tested; a few larger-scale points in the sparse groups could have checked whether the exponent transfers. The run thus repeats, with a flexible law, the pattern that makes cheapest-per-group the weakest non-agent rule.

%% file: sections/06_related_work.tex
\section{Related Work}
\label{sec:related}

\paragraph{AI for AI.}
\label{sec:related-ai}
AI systems already contribute to model development: they search over learning algorithms and update rules~\citep{real2020automlzero,chen2023lion}, optimize matrix multiplication and sorting~\citep{fawzi2022alphatensor,mankowitz2023alphadev}, evolve executable programs against measured objectives~\citep{romera2024funsearch,novikov2025alphaevolve}, and automate hypothesis generation, experimentation, and manuscript preparation~\citep{lu2024scientist,yamada2025scientistv2} as well as the assessment of research proposals~\citep{foster2026rpm}. Evaluating a proposal can itself require substantial computation. \bench{} targets the planning of that computation: which measurements an agent should acquire, and at what cost, to predict the behavior of future models.

\paragraph{Research-agent benchmarks.}
\label{sec:related-benchmarks}
Existing benchmarks evaluate machine-learning experimentation and engineering~\citep{huang2024mlagent,chan2025mle}, improvement of research objectives in executable environments~\citep{wijk2025re,nathani2025mlgym,garikaparthi2026researchgym,lupidi2026airs}, implementation fidelity and reproducibility~\citep{starace2025paper,siegel2025core}, and data analysis, hypothesis discovery, and scientific workflows~\citep{chen2024science,majumder2025discovery,bragg2026asta}. \bench{} makes the experimental evidence itself part of the agent's decision: its paired settings hold the prediction target fixed and vary only access to observations, and purchase records connect extrapolation quality to expenditure and coverage (\appref{app:related}).

\paragraph{Scaling laws and symbolic discovery.}
\label{sec:related-scaling}
Neural scaling laws relate performance to parameters, data, and compute and guide compute-optimal allocation~\citep{kaplan2020,hoffmann2022}; later work models transitions between regimes~\citep{caballero2023} and architectural dimensions beyond parameter count~\citep{mcleish2025}. Fitted exponents and extrapolations depend on the observed scale range, the data selected, and the fitting procedure~\citep{choshen2025,li2025misfitting}, and Active SLD places exactly this dependence under the agent's control. Symbolic regression searches for interpretable expressions that fit measured data~\citep{udrescu2020,cranmer2023}; SLDBench specializes this search to model-development experiments, and SLDAgent evolves the law expression together with its fitting program~\citep{lin2026}.

\paragraph{Cost-aware experimental design.}
\label{sec:related-cost}
Bayesian experimental design chooses measurements to reduce uncertainty~\citep{huan2013}, and Hyperband and BOHB allocate resources across configurations and fidelities~\citep{li2018hyperband,falkner2018bohb}. Closest to our setting, SL$^2$ selects experiments from finite pools with heterogeneous costs to fit supplied law families and extrapolate to expensive targets~\citep{li2026active}. In \bench{} the agent must also construct the law family, implement the fit, and decide when to stop, with cost reported rather than capped.

%% file: sections/07_limitations.tex
\section{Limitations and Conclusion}
\label{sec:limits}
\label{sec:conclusion}

\paragraph{Limitations.}
Replay reveals recorded outcomes, so an agent can buy only experiments that were run; it cannot propose new configurations, and the catalogue bounds what any policy can learn. Prices mix reported FLOPs, derived estimates, and proxies (\appref{app:corpus-cost}), so $\rho$ compares spending within a task but not absolute compute across tasks. The difficulty screen uses one reference law and thus favors tasks on which it fails. The evaluation covers \SLDEvaluatedSystems{} systems with three repeats on tasks from \SLDBenchmarkSources{} sources, so per-system estimates are coarse. Finally, cost is reported rather than traded off against quality, so systems are compared on two axes rather than ranked on one.

\paragraph{Conclusion.}
\bench{} turns scaling law discovery into an experimental-design problem: an agent buys priced experiments from real training records, constructs a law, and is scored on hidden larger-scale runs. Systems that are nearly indistinguishable when all data are given separate widely when they must buy it, and the loss comes from what they buy: agents favor cheap experiments, cheap purchase sets can leave conditions needed for extrapolation unobserved, and acquisition cost alone does not determine prediction quality. Non-agent rules indicate that, under a fixed law, variation in scale within each group matters more than reaching every group, and it is also what costs most. Agents that select experiments to discriminate between candidate laws, rather than by price, are a natural next step.

%% file: sections/10_statements.tex
\section*{Reproducibility statement}
Section~\ref{sec:benchmark} and the appendices specify the task construction, the reference law and difficulty screen, the execution environment and scoring, the statistical procedures, the purchase and coverage diagnostics, and the non-agent baselines. The accompanying scripts reproduce the corpus statistics, the compute accounting, and the non-agent baselines from the task files, and regenerate every experiment figure and table from the recorded evaluation results. Rerunning the agents additionally requires the released Harbor tasks and the models and agent harnesses listed in \tabref{tab:systems}.

\section*{AI use statement}
LLM agent (Codex) was used to polish the writing and figures and to assist with analysis scripts. The authors verified all text and figures and are fully responsible for the manuscript.

%% file: sections/appendix.tex
\section{Task inventory and evaluation protocol}
\label{app:inventory}

\subsection{Task inventory and construction}
\label{app:taskdetails}

The selected tasks contain \SLDBenchmarkObservations{} data points: \SLDBenchmarkPoolObservations{} candidates and \SLDBenchmarkTestObservations{} hidden test points (Table~\ref{tab:inventory}). Counts sum prediction records over the \SLDBenchmarkScenarios{} tasks and are not multiplied by settings, systems, or repeats; one training run or checkpoint may contribute several records.

\begin{table}[!htbp]
\centering\footnotesize
\caption{Data points of the \SLDBenchmarkScenarios{} selected tasks, \SLDBenchmarkObservations{} in total. Pool and test count the candidate and hidden test records; $G$ counts experimental groups and $d$ input features, as defined in Section~\ref{sec:formulation}.}
\label{tab:inventory}
\setlength{\tabcolsep}{3.5pt}
\begin{tabular}{llrrrr}
\toprule Code & Task & Pool & Test & $G$ & $d$\\\midrule
T01 & IBM BoolQ & 47 & 15 & 7 & 2 \\
T02 & IBM OpenBookQA & 47 & 15 & 7 & 2 \\
T03 & IBM OpenBookQA norm. & 47 & 15 & 7 & 2 \\
T04 & IBM RACE-high & 47 & 15 & 7 & 2 \\
T05 & IBM WinoGrande & 46 & 15 & 7 & 2 \\
T06 & OLMo COPA & 1,366 & 342 & 3 & 2 \\
T07 & OLMo PIQA norm. & 1,366 & 342 & 3 & 2 \\
T08 & OPT Code lines & 114 & 33 & 6 & 2 \\
T09 & OPT Human senses & 114 & 33 & 6 & 2 \\
T10 & OPT Implicatures & 114 & 33 & 6 & 2 \\
T11 & OPT Swahili proverbs & 114 & 33 & 6 & 2 \\
T12 & OPT Tao & 114 & 33 & 6 & 2 \\
T13 & Tokenization & 402 & 101 & 1 & 5 \\
T14 & Datablations / token--loss & 85 & 29 & 14 & 2 \\
T15 & DCLM controlled benchmark & 583 & 159 & 53 & 2 \\
T16 & Gemstones curve & 6,213 & 1,555 & 3 & 6 \\
T17 & MTEB / param.--performance & 575 & 161 & 23 & 1 \\
T18 & LAION / compute--downstream error & 32,160 & 8,417 & 837 & 2 \\
T19 & LAION / param.--downstream error & 25,210 & 9,591 & 5575 & 2 \\
T20 & LAION / sample--contrastive loss & 140 & 52 & 21 & 3 \\
T21 & LAION / sample--downstream error & 28,632 & 9,644 & 4119 & 2 \\
T22 & LAION / sample--segmentation error & 173 & 58 & 22 & 3 \\
T23 & Loss-to-loss transfer & 118,480 & 29,620 & 20 & 4 \\
T24 & Misfitting / checkpoint token--loss & 3,637 & 999 & 163 & 2 \\
T25 & Misfitting / final compute--loss & 190 & 50 & 5 & 3 \\
T26 & Overtraining loss & 664 & 168 & 24 & 3 \\
T27 & OWT / batch size--training loss & 12,012 & 3,003 & 1 & 3 \\
T28 & ProteinLM / compute--loss & 31,923 & 7,982 & 2 & 2 \\
T29 & Revisiting NMT learning curve & 40 & 13 & 5 & 2 \\
T30 & StepLaw hyperparam. loss & 1,528 & 383 & 1 & 4 \\
 & \textbf{Total} & \textbf{\SLDBenchmarkPoolObservations} & \textbf{\SLDBenchmarkTestObservations} & -- & -- \\
\bottomrule\end{tabular}
\end{table}

Records that share all input features and the group are merged into one record with their median target, so that no two records have identical inputs but different targets. Source-specific metadata defines the scale ordering and the construction of the test set, and \tabref{tab:sources} lists the upstream sources.

\begin{table}[!htbp]
\centering\footnotesize
\caption{Upstream sources of the selected tasks. Tasks from one source are not independent campaigns.}
\label{tab:sources}
\setlength{\tabcolsep}{3.5pt}
\begin{tabular}{p{.32\linewidth}p{.52\linewidth}r}
\toprule Source & Upstream project & Tasks\\\midrule
ColPret & \href{https://github.com/IBM/ColPret}{IBM/ColPret} & 12\\
LAION comparisons & \href{https://github.com/LAION-AI/scaling-laws-for-comparison}{LAION-AI/scaling-laws-for-comparison} & 5\\
Misfitting & \href{https://github.com/hadasah/scaling_laws}{hadasah/scaling\_laws} & 2\\
Compute-optimal tokenization & \href{https://github.com/facebookresearch/compute-optimal-tokenization}{facebookresearch/compute-optimal-tokenization} & 1\\
Datablations & \href{https://github.com/huggingface/datablations}{huggingface/datablations} & 1\\
DCLM & \href{https://github.com/mlfoundations/dclm}{mlfoundations/dclm} & 1\\
Gemstones & \href{https://github.com/mcleish7/gemstone-scaling-laws}{mcleish7/gemstone-scaling-laws} & 1\\
Portuguese MTEB & \href{https://huggingface.co/datasets/mteb-pt/mteb-pt-results}{mteb-pt/mteb-pt-results} & 1\\
Loss-to-loss transfer & \href{https://github.com/brendel-group/llm-line}{brendel-group/llm-line} & 1\\
Overtraining & \href{https://github.com/mlfoundations/scaling}{mlfoundations/scaling} & 1\\
OpenWebText histories & \href{https://github.com/shehper/scaling_laws}{shehper/scaling\_laws} & 1\\
ProteinLM compute--loss & \href{https://github.com/cxysteven/ScalingProteinLM}{cxysteven/ScalingProteinLM} & 1\\
Revisiting NMT & \href{https://github.com/google-research/google-research/tree/master/revisiting_neural_scaling_laws}{google-research/revisiting\_neural\_scaling\_laws} & 1\\
StepLaw & \href{https://github.com/step-law/steplaw}{step-law/steplaw} & 1\\
\bottomrule\end{tabular}
\end{table}

\subsection{Reference fitting and the difficulty screen}
\label{app:screening}

Collection and difficulty selection are separate stages. After structural curation the corpus contains \SLDStagingScenarios{} scenarios, of which \SLDActiveEligibleScenarios{} meet the eligibility conditions of Section~\ref{sec:tasks}; these conditions ensure that selecting experiments involves a genuine trade-off between information and cost.

The original SLDBench initializes SLDAgent with a power-law expression and the BFGS optimizer~\citep[Section~3]{lin2026}. The reference used for task selection instead fits a shared additive power-law family by bounded, multi-start nonlinear least squares with a soft-L1 penalty on the normalized residuals. Its numerically stabilized form for each group $g$ is
\begin{equation}
f_g(\mathbf x)=E_g+\sum_{j=1}^{d} A_{g,j}\exp\!\left[-\alpha_{g,j}\log(1+|x_j|)\right].
\label{eq:reference-law}
\end{equation}
Amplitudes and exponents may take either sign, which accommodates both decreasing losses and increasing scores. The fit uses the candidates alone, falls back to a linearized or pooled fit for groups with too few points, and clips predictions to the observed range of the training targets.

The Active reference buys the entire pool and uses the same implementation. Because numerical fits under a time limit need not coincide, the screen requires both scores to satisfy
\begin{equation}
\max\!\left\{R^2_{\mathrm{ref,Passive}},\;R^2_{\mathrm{ref,Active}}\right\}
\leq \SLDDifficultyThreshold.
\label{eq:difficulty-screen}
\end{equation}
This retains \SLDBenchmarkScenarios{} scenarios, and no agent score enters the selection. Difficulty is defined relative to this fixed reference law, so the selected tasks are not a representative sample of the corpus.

\subsection{Evaluation protocol and acquisition accounting}
\label{app:protocol}

Both settings of a task use the same hidden test file, and the catalogue order is shuffled. Agents submit \texttt{law.py}, \texttt{explain.md}, and any required parameter files; the law maps a dictionary of inputs and a group identifier to a prediction. Only the verifier accesses the test targets.

Claude Opus 5.5 uses Claude Code 2.1.280 with model identifier \texttt{claude-opus-5-5[1m]}, a 1,000,000-token context window, and high reasoning effort. The agent and verifier time limits are 3,600 and 600 seconds, respectively.

For every task, both versions contain the same multiset of features, groups, and targets; Passive files additionally expose task-specific columns. Twelve tasks derive from ColPret, five from LAION, and two from Misfitting, which motivates the source-level bootstrap of \appref{app:score-rules}.

Acquisition cost is the sum of catalogue prices over the distinct purchased candidates, normalized within each task; repeated queries incur no additional charge. Portuguese MTEB is priced by parameter count because its FLOPs are unavailable. The price of a checkpoint may be the cumulative cost of its training run, and the acquisition cost does not deduct computation shared between checkpoints.

\subsection{Source-unit compute coverage}
\label{app:corpus-cost}

Of the selected catalogues, \SLDExplicitPriceScenarios{} use a stored price column and \SLDProxyPriceScenarios{} construct prices as $6ND$. A stored column does not determine the provenance of a cost, since a column named \texttt{cost\_flops} can contain an estimate or a proxy. We distinguish C3 source-reported FLOPs, C2 FLOPs derived under documented assumptions, C1 non-FLOP proxies, and C0 unavailable costs. In particular, Portuguese MTEB's \texttt{cost\_flops} equals the parameter count and is therefore C1, whereas OpenWebText's stored training-compute column is derived from $6ND$ and is C2 rather than C3.

To quantify the training computation that the tasks represent, we aggregate over source units, i.e., training runs or configurations, rather than over prediction records, and compute
\begin{equation}
C_{\mathrm{tier}}=\sum_{u\in\mathcal U_{\mathrm{tier}}}\max_{r\in u} c_r.
\label{eq:source-cost}
\end{equation}
Pool and test are joined before grouping, and data shared by the two settings are counted once. Each training trajectory is thus charged once, even if it contributes several checkpoints or metrics. For tokenization, the largest reported budget of each configuration gives a lower bound on its compute without assuming that each reported budget corresponds to an independent training run.

\begin{table}[!htbp]
\centering\small
\caption{Training compute represented by the selected tasks. Units are training runs or configurations, and totals take the maximum recorded cost of each unit over pool and test. C3 and C2 totals are kept separate.}
\label{tab:source-cost}
\setlength{\tabcolsep}{5pt}
\begin{tabular}{lllrr}
\toprule Source & Tier & Accounting unit & Units & FLOPs\\\midrule
Compute-optimal tokenization & C3 & Configuration & \SLDTokenizationUnits & $\SLDTokenizationFLOPs$\\
ProteinLM & C3 & Source run & \SLDProteinUnits & $\SLDProteinFLOPs$\\
OpenWebText & C2 & Source run & \SLDOWTUnits & $\SLDOWTFLOPs$\\
\bottomrule
\end{tabular}
\end{table}

The \SLDReportedCostCoverage{} C3 sources alone give the lower bound of $\SLDBenchmarkReportedFLOPs$ FLOPs stated in the introduction. The C2 estimate in Table~\ref{tab:source-cost}, all C1 proxies, and sources without a verified mapping from units to costs are excluded, so the bound is neither a corpus total nor the expenditure of an agent.

\section{Evaluation results and statistical methods}
\label{app:scores}

\subsection{Score inclusion and aggregation}
\label{app:score-rules}

Medians and error ratios use finite scores. Table~\ref{tab:coverage-v4} reports, for each system and setting, the percentage of all scheduled runs, including unscored ones, whose score exceeds each threshold.

For setting comparisons, the effect is $\log_{10}[(1-R_A^2)/(1-R_P^2)]$; positive values favor Passive. An error floor of $10^{-12}$ handles zero error. Active and Passive runs are paired by system, task, and repeat index. We take the median effect within each of the 14 sources, then report the median across sources with a 95\% percentile interval from 10{,}000 bootstrap resamples of sources.

\begin{table}[!htbp]
\centering\small
\caption{Percentage of all scheduled runs whose $R^2$ exceeds each threshold.}
\label{tab:coverage-v4}
\setlength{\tabcolsep}{3.5pt}
\begin{tabular}{lrrrrrr}
\toprule System & \shortstack{Passive\\> 0} & \shortstack{Active\\> 0} & \shortstack{Passive\\> .5} & \shortstack{Active\\> .5} & \shortstack{Passive\\> .9} & \shortstack{Active\\> .9}\\\midrule
Claude Opus 5.5 & 96.7\% & 77.8\% & 80.0\% & 55.6\% & 26.7\% & 15.6\% \\
DeepSeek V4 Pro & 86.7\% & 78.9\% & 64.4\% & 51.1\% & 15.6\% & 11.1\% \\
DeepSeek V4 Flash & 84.4\% & 64.4\% & 66.7\% & 41.1\% & 11.1\% & 5.6\% \\
GPT-5.6 Sol & 85.6\% & 71.1\% & 70.0\% & 51.1\% & 20.0\% & 12.2\% \\
Gemini 3.8 Flash & 85.6\% & 62.2\% & 70.0\% & 42.2\% & 18.9\% & 5.6\% \\
Qwen3.8 Flash & 92.2\% & 65.6\% & 74.4\% & 42.2\% & 20.0\% & 3.3\% \\
Hy4-Preview & 94.4\% & 60.0\% & 80.0\% & 42.2\% & 23.3\% & 3.3\% \\
Kimi K3 & 83.3\% & 66.7\% & 72.2\% & 45.6\% & 18.9\% & 8.9\% \\
\bottomrule\end{tabular}
\end{table}

\subsection{Best-of-three prediction quality}
\label{app:best-at-three}

For each system, task, and setting, best@3 takes the highest held-out $R^2$ among the three repeats. Table~\ref{tab:best-at-three} reports the median of these 30 task-wise maxima, computed separately for Passive and Active:
\begin{equation}
\operatorname{best@3}(s)
=\operatorname{median}_{t=1,\ldots,30}\;\max_{r\in\{1,2,3\}} R^2_{s,t,r},
\qquad s\in\{\mathrm{Passive},\mathrm{Active}\}.
\end{equation}
This complements the run-level medians in Table~\ref{tab:systems} by summarizing the best outcome obtained for each task across three attempts.

\begin{table}[!htbp]
\centering\small
\caption{Best@3 $R^2$: median over 30 tasks of the highest score across three repeats per task and setting. Bold marks the best value in each column.}
\label{tab:best-at-three}
\setlength{\tabcolsep}{12pt}
\begin{tabular}{lrr}
\toprule
Model & Passive best@3 $R^2$ & Active best@3 $R^2$\\
\midrule
Claude Opus 5.5 & \textbf{0.8179} & 0.6877 \\
Kimi K3 & 0.7684 & 0.6919 \\
DeepSeek V4 Pro & 0.7545 & \textbf{0.7583} \\
GPT-5.6 Sol & 0.7744 & 0.7438 \\
Qwen3.8 Flash & 0.7978 & 0.6673 \\
Hy4-Preview & 0.7970 & 0.5659 \\
Gemini 3.8 Flash & 0.7895 & 0.5611 \\
DeepSeek V4 Flash & 0.7494 & 0.5758 \\
\bottomrule
\end{tabular}
\end{table}

\subsection{Non-agent acquisition baselines}
\label{app:baselines}

The baselines ask how well the fixed reference law of \appref{app:screening} extrapolates from a purchase set chosen without looking at any outcome. For each task and each budget $\rho$ of 0.5, 1, 2, 5, 10, 20, 50, and 100\%, a rule scans its candidate order and buys each candidate whose price still fits in the remaining budget $\rho\,C(\mathcal U)$. Cheapest-first orders candidates by price. Random uses three fixed permutations, and a task's random score is the median over the three seeds, with failed fits ranked worst. Cheapest-per-group orders the groups by their cheapest candidate and visits them round-robin, taking each group's next-cheapest candidate. Price-spread visits the groups in the same order but ranks each group's candidates by price and takes them coarse to fine: cheapest, most expensive, middle, quartiles, and so on. At $\rho=100\%$ every rule buys the full pool, so it is fitted once. Each fit runs the task's reference implementation on the purchased subset and is scored on the hidden test set with the same unclipped $R^2$ that the verifier computes for agent submissions.

Table~\ref{tab:baselines} lists each rule's median $R^2$ and the share of tasks on which it beats cheapest-first at the same budget. Because all rules share one law, differences among them isolate the effect of acquisition, whereas comparisons with the agents (Section~\ref{sec:value}) also reflect law construction.

\begin{table}[!htbp]
\centering\footnotesize
\caption{Non-agent acquisition baselines under the fixed reference law. Median $R^2$ is over the 30 tasks, all of which have finite scores. ``Beats cheapest-first'' is the percentage of the 30 tasks on which a rule scores a strictly higher $R^2$ than cheapest-first at the same budget, so ties count against the rule.}
\label{tab:baselines}
\setlength{\tabcolsep}{3.5pt}
\begin{tabular}{rrrrrrrr}
\toprule
& \multicolumn{4}{c}{Median $R^2$} & \multicolumn{3}{c}{Beats cheapest-first (\%)}\\
\cmidrule(lr){2-5}\cmidrule(l){6-8}
$\rho$ (\%) & cheapest & random & per group & spread & random & per group & spread\\
\midrule
0.5 & $-$1.104 & $-$0.312 & $-$2.541 & $-$0.533 & 66.7 & 23.3 & 66.7 \\
1 & $-$0.646 & $-$0.297 & $-$2.029 & $-$0.488 & 63.3 & 23.3 & 63.3 \\
2 & $-$0.811 & $-$0.082 & $-$1.456 & $-$0.005 & 73.3 & 16.7 & 76.7 \\
5 & $-$0.182 & $-$0.003 & $-$0.664 & 0.063 & 73.3 & 46.7 & 70.0 \\
10 & 0.065 & 0.085 & $-$0.044 & 0.027 & 60.0 & 30.0 & 50.0 \\
20 & 0.071 & 0.150 & $-$0.276 & 0.173 & 53.3 & 16.7 & 60.0 \\
50 & 0.272 & 0.360 & 0.138 & 0.323 & 66.7 & 23.3 & 63.3 \\
100 & 0.455 & 0.455 & 0.455 & 0.455 & -- & -- & -- \\
\bottomrule\end{tabular}
\end{table}

\section{Purchase behavior and system-specific patterns}
\label{app:behavior}

Purchase statistics use runs with at least one purchase, and those involving prediction quality also require a finite score. A full-pool run buys every candidate; a partial run buys a strict subset.

Price preference is measured by the ratio of the mean purchased price to the catalogue mean, which is one for full-pool runs. Batch statistics count newly purchased candidates. In the first and last batches, each candidate receives the midrank of its price within its group, scaled to $[0,1]$ with lower values cheaper; we average these positions within each batch and report the median over multi-batch runs. Table~\ref{tab:buying-v4} reports the full-pool share and the medians of point share, cost share, and price ratio, and Table~\ref{tab:system-behavior} the spread of spending and the stability of outcomes across repeats.

\begin{table}[!htbp]
\centering\footnotesize
\caption{Purchase behavior. Full pool is the percentage of runs that buy every candidate; the other columns are separate medians over runs, and those of batch price positions over multi-batch runs (lower is cheaper).}
\label{tab:buying-v4}
\setlength{\tabcolsep}{2.5pt}
\begin{tabular}{lrrrrrr}
\toprule System & Full pool & Point share & Cost share & Price ratio & \shortstack{First-batch\\price position} & \shortstack{Last-batch\\price position}\\\midrule
Claude Opus 5.5 & 1.1\% & 21.5\% & 4.01\% & 0.237 & 0.197 & 0.608 \\
DeepSeek V4 Pro & 7.8\% & 53.0\% & 21.51\% & 0.586 & 0.347 & 0.512 \\
DeepSeek V4 Flash & 26.7\% & 28.6\% & 17.59\% & 0.904 & 0.287 & 0.500 \\
GPT-5.6 Sol & 11.1\% & 69.5\% & 35.20\% & 0.716 & 0.442 & 0.500 \\
Gemini 3.8 Flash & 15.6\% & 47.8\% & 7.78\% & 0.271 & 0.126 & 0.375 \\
Qwen3.8 Flash & 2.2\% & 34.0\% & 5.35\% & 0.248 & 0.309 & 0.414 \\
Hy4-Preview & 0.0\% & 40.4\% & 3.44\% & 0.182 & 0.253 & 0.374 \\
Kimi K3 & 10.0\% & 43.5\% & 17.75\% & 0.595 & 0.269 & 0.500 \\
\bottomrule\end{tabular}
\end{table}

\begin{table}[!htbp]
\centering\footnotesize
\caption{Spending spread and repeat stability. Costs are percentages of the full-pool cost and the next three columns percentages of Active runs; sign changes of $R^2$ and more than tenfold error ranges across the three repeats are percentages of the tasks with three finite Active repeats.}
\label{tab:system-behavior}
\setlength{\tabcolsep}{2.5pt}
\begin{tabular}{lrrrrrr}
\toprule System & Cost median (\%) & Cost ${<}1\%$ & Cost ${>}90\%$ & Full pool & \shortstack{Sign\\changes} & ${>}10\times$ error\\\midrule
Claude Opus 5.5 & 4.01 & 23.3\% & 1.1\% & 1.1\% & 20.0\% & 6.7\% \\
DeepSeek V4 Pro & 21.51 & 12.2\% & 8.9\% & 7.8\% & 26.7\% & 23.3\% \\
DeepSeek V4 Flash & 17.59 & 22.2\% & 27.8\% & 26.7\% & 33.3\% & 23.3\% \\
GPT-5.6 Sol & 35.20 & 7.8\% & 13.3\% & 11.1\% & 43.3\% & 33.3\% \\
Gemini 3.8 Flash & 7.78 & 21.1\% & 15.6\% & 15.6\% & 46.7\% & 20.0\% \\
Qwen3.8 Flash & 5.35 & 23.3\% & 2.2\% & 2.2\% & 46.7\% & 10.0\% \\
Hy4-Preview & 3.44 & 31.1\% & 0.0\% & 0.0\% & 37.0\% & 11.1\% \\
Kimi K3 & 17.75 & 15.6\% & 13.3\% & 10.0\% & 37.5\% & 16.7\% \\
\bottomrule\end{tabular}
\end{table}

\section{Coverage and extrapolation case studies}
\label{app:cases}

Point coverage is the fraction of candidates purchased; group coverage is the fraction of groups with at least one purchase. Experimental groups describe within-task contexts rather than data sources and need not enumerate all parameter combinations. Condition coverage examines parameter values, their combinations, and observed scale ranges through task-specific checks.

Loss of independent parameter variation means that acquired observations distinguish fewer parameter contributions than the full catalogue under the same input representation. Technically, we compare design-matrix rank within each group. Inputs use $\operatorname{sign}(x)\log(1+|x|)$, with a constant column; varying columns and standardization are fixed from the full catalogue. Existing catalogue collinearity is not counted as an acquisition loss, while entirely unobserved groups count as losses. Table~\ref{tab:geometry-v4} separately reports missing groups and runs with reduced independent parameter variation.

\begin{table}[!htbp]
\centering\footnotesize
\caption{Experimental coverage, as percentages of all Active runs of each system. Independent variation is measured under the fixed input representation defined above; a run that observes every group can still leave fewer parameter contributions distinguishable than the full catalogue.}
\label{tab:geometry-v4}
\setlength{\tabcolsep}{2.5pt}
\begin{tabular}{lrrrr}
\toprule System & All groups & Missing groups & \shortstack{Independent\\variation lost} & \shortstack{All groups;\\variation lost}\\\midrule
Claude Opus 5.5 & 80.0\% & 20.0\% & 46.7\% & 26.7\% \\
DeepSeek V4 Pro & 88.9\% & 11.1\% & 24.4\% & 13.3\% \\
DeepSeek V4 Flash & 83.3\% & 16.7\% & 28.9\% & 12.2\% \\
GPT-5.6 Sol & 100.0\% & 0.0\% & 3.3\% & 3.3\% \\
Gemini 3.8 Flash & 96.7\% & 3.3\% & 14.4\% & 11.1\% \\
Qwen3.8 Flash & 86.7\% & 13.3\% & 34.4\% & 21.1\% \\
Hy4-Preview & 88.9\% & 11.1\% & 47.8\% & 36.7\% \\
Kimi K3 & 90.0\% & 10.0\% & 18.9\% & 8.9\% \\
\bottomrule\end{tabular}
\end{table}

The OpenWebText example has one explicit group but 15 batch-size and learning-rate combinations. Hy4-Preview acquires 51 observations covering 13 combinations; 30 observations use batch size 2. The combinations of learning rate $10^{-4}$ with batch size 64 or 128 are never purchased, although both occur in the catalogue and test set; Table~\ref{tab:condition-omissions} lists all such omissions.

\begin{table}[!htbp]
\centering\small
\caption{OpenWebText parameter combinations left unobserved by 12.5\% of the runs of this task.}
\label{tab:condition-omissions}
\setlength{\tabcolsep}{3.5pt}
\begin{tabular}{llrrrr}
\toprule System & Run & Points & Batch & Learning rate & $R^2$\\\midrule
Hy4-Preview & A & 44 & 4 & 0.0001 & $-$2.392 \\
Hy4-Preview & A & 44 & 64 & 0.0001 & $-$2.392 \\
Hy4-Preview & B & 51 & 64 & 0.0001 & $-$83.305 \\
Hy4-Preview & B & 51 & 128 & 0.0001 & $-$83.305 \\
Kimi K3 & A & 135 & 16 & 0.0001 & $-$10.580 \\
\bottomrule\end{tabular}
\end{table}

In the T22 case of Section~\ref{sec:value}, DeepSeek V4 Pro submits $\hat y_g(D)=E_g+A_gD^{-0.05}$, with the exponent shared by all groups. It buys all eight candidates of four inexpensive groups, two candidates in each of 14 groups, and three to six in each of the remaining four (\tabref{tab:case-v4}).

\begin{table}[!htbp]
\centering\footnotesize
\caption{All 22 groups of the T22 case, sorted by share of the squared test error (SSE); $D$ in millions.}
\label{tab:case-v4}
\setlength{\tabcolsep}{3.5pt}\renewcommand{\arraystretch}{0.92}
\begin{tabular}{lrrrrr}
\toprule Group & Bought/pool & Bought D & Test D & RMSE & SSE share\\\midrule
ViT-B-16\_mammut & 2/8 & 1.28--3.072 & 640--3072 & 0.2867 & 28.16\% \\
ViT-B-32\_mammut & 2/8 & 1.28--3.072 & 640--3072 & 0.2826 & 27.35\% \\
ViT-B-16-text-plus\_clip & 2/8 & 1.28--3.072 & 640--3072 & 0.2273 & 17.70\% \\
ViT-S-14\_mammut & 2/8 & 1.28--3.072 & 640--3072 & 0.1469 & 7.40\% \\
ViT-M-32\_mammut & 2/8 & 1.28--3.072 & 640--3072 & 0.1319 & 5.96\% \\
ViT-S-14\_clip & 2/7 & 1.28--3.072 & 1280--3072 & 0.1083 & 2.68\% \\
ViT-B-14\_clip & 2/6 & 1.28--3.072 & 1280--3072 & 0.1005 & 2.31\% \\
ViT-L-32\_mammut & 2/8 & 1.28--3.072 & 1280--3072 & 0.0726 & 1.20\% \\
ViT-S-32\_mammut & 8/8 & 1.28--307.2 & 640--3072 & 0.0584 & 1.17\% \\
ViT-S-16\_mammut & 2/8 & 1.28--3.072 & 640--3072 & 0.0542 & 1.00\% \\
ViT-M-16\_clip & 2/8 & 1.28--3.072 & 640--3072 & 0.0536 & 0.98\% \\
ViT-M-16\_mammut & 2/8 & 1.28--3.072 & 640--3072 & 0.0522 & 0.93\% \\
ViT-L-14\_mammut & 3/8 & 1.28--6.4 & 640--3072 & 0.0504 & 0.87\% \\
ViT-L-32\_clip & 5/8 & 1.28--30.72 & 640--3072 & 0.0485 & 0.80\% \\
ViT-L-16\_mammut & 2/8 & 1.28--3.072 & 1280--3072 & 0.0481 & 0.53\% \\
ViT-M-32\_clip & 8/8 & 1.28--307.2 & 640--3072 & 0.0288 & 0.28\% \\
ViT-S-32\_clip & 8/8 & 1.28--307.2 & 640--3072 & 0.0239 & 0.20\% \\
ViT-L-16\_clip & 2/8 & 1.28--3.072 & 640--3072 & 0.0275 & 0.17\% \\
ViT-S-16\_clip & 8/8 & 1.28--307.2 & 1280--3072 & 0.0265 & 0.16\% \\
ViT-L-14\_clip & 4/8 & 1.28--12.8 & 640--3072 & 0.0170 & 0.10\% \\
ViT-B-14\_mammut & 2/8 & 1.28--3.072 & 640--1280 & 0.0098 & 0.02\% \\
ViT-B-32\_clip & 6/8 & 1.28--64 & 1280--3072 & 0.0083 & 0.02\% \\
\bottomrule\end{tabular}
\end{table}

\section{Comparison with research benchmarks}
\label{app:related}

Table~\ref{tab:benchmarks} summarizes the benchmarks discussed in Section~\ref{sec:related}. SLDBench evaluates law construction from supplied observations and SL$^2$ evaluates acquisition under specified law families; \mbox{\bench} evaluates acquisition and law construction jointly, under a priced catalogue.

\begin{table}[!htbp]
\centering\footnotesize
\caption{Benchmarks discussed in Section~\ref{sec:related}, summarized by their primary evaluated activity and outcome.}
\label{tab:benchmarks}
\setlength{\tabcolsep}{4pt}\renewcommand{\arraystretch}{0.92}
\begin{tabular*}{\linewidth}{@{\extracolsep{\fill}}lll@{}}
\toprule Benchmark & Evaluated activity & Principal outcome\\\midrule
MLAgentBench / MLE-bench & ML experimentation and engineering & Model or competition score \\
MLGym & Open-ended AI research tasks & Task-specific improvement \\
RE-Bench & Research engineering & Progress relative to experts \\
ResearchGym & Rediscovering a withheld method & Gain over supplied baselines \\
AIRS-Bench & Research without baseline code & Metrics versus published SOTA \\
PaperBench / CORE-Bench & Paper replication / reproduction & Rubric score / reproduction \\
ScienceAgentBench & Data-driven scientific programs & Program execution outcomes \\
DiscoveryBench & Data-driven hypothesis discovery & Hypothesis facets \\
AstaBench & Scientific research assistance & Suite-level task outcomes \\
SLDBench & Law discovery from completed runs & Predictive law quality \\
SL$^2$ & Acquisition for supplied law families & Extrapolation under cost budgets \\
SLDBench v2 (ours) & Priced acquisition and law discovery & Prediction, cost, and coverage \\
\bottomrule\end{tabular*}
\end{table}